\documentclass[conference]{IEEEtran}
\IEEEoverridecommandlockouts

\usepackage{cite}
\usepackage{amsmath,amssymb,amsfonts}
\usepackage{algorithmic}
\usepackage{algorithm}
\usepackage{graphicx}
\usepackage{stmaryrd}
\usepackage[dvipsnames]{xcolor}
\usepackage{array}
\usepackage{commath}
\usepackage{sidecap}
\usepackage{stfloats}
\usepackage{tabularx, boldline}
\usepackage{rotating,booktabs,multirow}
\usepackage{mathtools}
\usepackage{flexisym}
\usepackage{breqn}
\usepackage{url}
\ifCLASSOPTIONcompsoc
    \usepackage[caption=false, font=normalsize, labelfont=sf, textfont=sf]{subfig}
\else
\usepackage[caption=false, font=footnotesize]{subfig}
\fi
\newcommand{\mkv}{-\!\!\!\!\minuso\!\!\!\!-}

\usepackage{authblk}
\def\BibTeX{{\rm B\kern-.05em{\sc i\kern-.025em b}\kern-.08em
    T\kern-.1667em\lower.7ex\hbox{E}\kern-.125emX}}

\makeatletter
\newcommand\fs@betterruled{%
  \def\@fs@cfont{\bfseries}\let\@fs@capt\floatc@ruled
  \def\@fs@pre{\vspace*{0.1in}\hrule height.8pt depth0pt \kern2pt}%
  \def\@fs@post{\kern2pt\hrule\relax}%
  \def\@fs@mid{\kern2pt\hrule\kern2pt}%
  \let\@fs@iftopcapt\iftrue}
\floatstyle{betterruled}
\restylefloat{algorithm}
\makeatother

\begin{document}

\title{$\alpha$-Mutual Information: A Tunable Privacy Measure for Privacy Protection in Data Sharing\vspace{-0.6ex}}
\author[1]{\vspace{-0.3ex}MirHamed Jafarzadeh~Asl}
\author[2]{Mohammadhadi~Shateri}
\author[1]{Fabrice~Labeau\vspace{-2mm}}
\affil[1]{\small Department of Electrical and Computer Engineering, McGill University, QC, Canada, \protect\\ Email: mirhamed.jafarzadehasl@mail.mcgill.ca, fabrice.labeau@mcgill.ca}
\affil[2]{Department of Systems Engineering, École de Technologie Supérieure, QC, Canada, \protect\\ Email: mohammadhadi.shateri@etsmtl.ca
\vspace{-3.4ex}
}

\maketitle
\thispagestyle{plain}
\pagestyle{plain}


\begin{abstract}
This paper adopts Arimoto's $\alpha$-Mutual Information as a tunable privacy measure, in a privacy-preserving data release setting that aims to prevent disclosing private data to adversaries. By fine-tuning the privacy metric, we demonstrate that our approach yields superior models that effectively thwart attackers across various performance dimensions. We formulate a general distortion-based mechanism that manipulates the original data to offer privacy protection. The distortion metrics are determined according to the data structure of a specific experiment. We confront the problem expressed in the formulation by employing a general adversarial deep learning framework that consists of a releaser and an adversary, trained with opposite goals. This study conducts empirical experiments on images and time-series data to verify the functionality of $\alpha$-Mutual Information. We evaluate the privacy-utility trade-off of customized models and compare them to mutual information as the baseline measure. Finally, we analyze the consequence of an attacker's access to side information about private data and witness that adapting the privacy measure results in a more refined model than the state-of-the-art in terms of resiliency against side information.
\end{abstract}


\vspace{-2ex}
\section{Introduction}

\vspace{-0.5ex}
Despite technological advancements and increased data generation, the need for data sharing has risen dramatically.
However, data sharing always carries the risk of security breaches, with unauthorized entities trying to extract private information from shared data.
Notably, the privacy problem in data sharing differs from the data security issue. In data release privacy, any authorized receiver of the data is considered an anticipated invader. Therefore, data security methods are unprofitable in data sharing~\cite{SM_data_security_vs_data_sharing}.
As data sharing has progressed with advancements in speed, feasibility, etc., addressing various privacy issues has become more challenging than ever before. For instance,
many social media applications require individuals to share private data online~\cite{Dynamics_in_Data_Privacy_and_Sharing_Economics}.
Hence, various privacy-protecting techniques for data sharing have been studied for years.
Differential Privacy (DP) has received significant attention in this area, especially due to its low computational overhead~\cite{dp_2_2}. Although DP prioritizes data privacy, it may not be ideal for applications where preserving the utility of shared data is crucial, as it does not specifically address other data properties~\cite{dp_5_utility_issue}.

\vspace{-0.6ex}
\subsection{Related work}
Considering the mentioned shortcoming of DP, information-theoretical approaches are widely applied in privacy protection, offering improved privacy-utility trade-offs (PUTs)~\cite{hadi-journal-alpha1-ECOwoSI, MI__Privacy-Preserving_Representation_Learning_on_Graphs, MI__DI__hadi_deep-directed}.
Mutual Information (MI) has been popular in information-theoretical privacy measures.
In \cite{MI__Privacy-Preserving_Representation_Learning_on_Graphs}, an MI-based method is designed to prevent leakage of private features in representation learning methods on graphs.
Besides, efforts are made to extract the most from the patterns in data to determine convenient metrics. One such example is demonstrated in \cite{MI__DI__hadi_deep-directed}, where Directed Information (DI) is selected as the privacy measure. Nonetheless, in order to achieve flexible PUTs, the necessity of discovering a tunable privacy measure has been perceived. An adjustable metric allows for tailoring the privacy definition to specific use cases, enhancing performance, and demonstrating the capacity of information-theoretical strategies.

Configurable measures of information leakage based on R{\'e}nyi entropy~\cite{renyi1961measures} and Arimoto $\alpha$-mutual information ($\alpha$-MI)~\cite{ARIMOTO} are designed in the literature. Suggesting tunable metrics in~\cite{first_paper_A_Tunable_Measure_for_Information_Leakage}, authors introduce $\alpha$-leakage as a measure of information disclosure that quantifies how much an adversary can infer a specific private attribute of the data. The definitions have been extended in~\cite{alpha-beta-leakage}.
To the best of our knowledge, the closest study to our work is presented in \cite{Generating_Fair_Universal_Representations_Using_Adversarial_Models}, which employs $\alpha$-loss (equivalent to using Arimoto $\alpha$-MI as privacy measure) within an adversarial learning framework for data sharing.
However, they formulated the problem as a minimax game with constraints, which has been demonstrated to be unstable with regard to loss in deep learning \cite{gan}. Moreover, the influence of the $\alpha$ parameter in such a tunable measure and its impact on improving PUT has not been investigated.

Furthermore, one may assess privacy-preserving data-sharing systems regarding their effectiveness in a scenario where a malicious attacker has access to sort of side information (SI) correlated with private data. The authors in \cite{side-info-hadi} analyze this problem. However, we show that customizing the privacy measure can lead to more reliable models than in~\cite{side-info-hadi} in terms of PUT. Notably, the robustness of Maximal $\alpha$-leakage to arbitrary SI is studied in~\cite{Robustness-of-Maximal-alpha-Leakage-to-Side-Information}; however, their conclusion is drawn based on the availability of ground truth private attributes. Although this notion is reasonable in the training phase of a framework, it is unrealistic to imagine that private features are known in the testing stage. Moreover, the assumption of having all attributes of the original data as private features might not be practical in many applications.

\vspace{-0.6ex}
\subsection{Contributions}
\vspace{-0.4ex}
In this paper, a tunable privacy measure has been adopted on distortion-based privacy-preserving data release models. The main contributions of this work are as follows:

\begin{enumerate}
    \item To the best of our knowledge, this is the first time that the impacts of the $\alpha$ parameter are practically investigated in $\alpha$-MI as a measure of privacy in the privacy-preserving data release.
    \item The impact of the tunable privacy measure is illustrated in the presence of SI that is correlated with the sensitive information of shareable data.
    \item We suggest a framework that uses a stable strategy to address the optimization problem of privacy-preserving data release as opposed to a minimax formulation~\cite{Generating_Fair_Universal_Representations_Using_Adversarial_Models}.
    \item Our framework is customized for several datasets with different structures to examine the advantages of using an adaptable privacy measure.
\end{enumerate}


\vspace{-1ex}
\subsection*{Notation and conventions}
A sequence of random variables $\hspace{-0.7mm}(X_1,\hspace{-0.2mm} X_2,\hspace{-0.2mm} \dots,\hspace{-0.2mm} X_T)$ is shown as $X^T$. A sample batch from $X^T$ is written as $\{x^{(b)T}\}^{B}_{b=1}$. The probability distribution of $X_t$ is $p_{\hspace{-0.3mm}X_t}$, and the conditional distribution of $X_t$ given $Y_t$ is shown as $p_{\hspace{-0.3mm}X_t\hspace{-0.2mm}|\hspace{-0.2mm}Y_t}$. The conditional distribution $X^T$ given $Y^T$ would be $p_{\hspace{-0.3mm}X^T\hspace{-0.2mm}|\hspace{-0.2mm}Y^T}\hspace{-0.2mm}$.
A Markov chain composed of $X,\hspace{-0.5mm}Y\hspace{-0.5mm},\hspace{-0.5mm}$ and $\hspace{-0.2mm}Z$ is written as $X \hspace{-0.7mm} \mkv \hspace{-0.7mm} Y \hspace{-1mm} \mkv \hspace{-1mm} Z$.
The expectation of a function $f$ with respect to $p_{\hspace{-0.3mm}X}$ is denoted as $E[f\hspace{-1mm}\left(\hspace{-0.4mm}X\hspace{-0.2mm}\right)]$.
The Kullback-Leibler~(KL) divergence between two distributions $p_1$ and $p_2$ is represented as $\text{KL}(p_1||p_2)$.



\section{Problem Formulation and Training Objective}
\label{sec:formulation}
\vspace{-0.2ex}
Let variables $Y^T$ denote the users' useful data. This data may be metered power consumption of houses over $T$ time slots, or any non-sequence data ($T=1$) such as patients' health conditions. Private variables $X^T$ represent the sensitive information that a particular user is unwilling to share in public, e.g., people's identities in the data collected by social media. We also define observed variables $W^T$ as the variables that would normally be released or shared. We assume that $W^T$ is not independent of $X^T$. The private information $X^T$ may be present, together with the $Y^T$, in $W^T$, or $X^T$ is correlated with $Y^T$ and $W^T$ is formed of $Y^T$. Therefore, for a particular task, sensitive information should be eliminated from valuable data before sharing the data publicly. In this scenario, a privacy-preserving system is of interest. This system contains a releaser that creates a new representation of $Y^T$, denoted as $Z^T$, generated by distorting $Y^T$ to follow two objectives simultaneously: the releaser aims to hide private data from any possible attacker interested in inferring them from released data; at the same time, it tries to preserve useful data, as much as possible, based on specific criteria. Therefore, measures are needed to quantify the released data's privacy performance and utility achievement (i.e., preserving useful attributes).
Moreover, harmful attackers could have access to some supplementary (side) information, $S$, that can assist them in attaining higher inference performance.
To quantify the distortion between $Z^T$ and $Y^T$, we define a distortion measure as $\mathcal{D}(Z^T,Y^T) \triangleq \mathbb{E}[d(Z^T,Y^T)]$, where
$d\hspace{-0.3mm}:\hspace{-0.5mm} \mathbb{R}^T\hspace{-0.7mm}\times \hspace{-0.1mm}\mathbb{R}^T \hspace{-0.7mm}\rightarrow\hspace{-0.4mm} \mathbb{R}$ can be any distortion metric on $\mathbb{R}^T$.
Here, Arimoto's 
$\alpha$-Mutual Information is proposed for the privacy measure in the releaser as $I^{A}_{\alpha}(X; Z) = H_{\alpha}(X) - H^{A}_{\alpha}({X}|Z)$~\cite{ARIMOTO}, where
$H_{\alpha}(X)$
is the R{\'e}nyi entropy of order $\alpha \in (0,1) \cup (1,\infty)$~\cite{renyi1961measures} written as:
\vspace{-1ex}
\begin{equation}
    H_{\alpha}(X) \hspace{-0.5mm}= \hspace{-0.5mm}\frac{\alpha}{1-\alpha} \log \hspace{-0.5mm} \left(\sum_{x}{\hspace{-0.5mm}p^{\alpha}_{X}(x)}\right)^{\hspace{-1.5mm}\frac{1}{\alpha}}\hspace{-1mm} = \frac{\alpha}{1-\alpha} \log\hspace{-0.5mm} {\lVert p_{X} \rVert}_{\alpha},\vspace{-0.3ex}
\end{equation}
and $\hspace{-0.4mm}H^{A}_{\alpha}\hspace{-0.3mm}(\hspace{-0.3mm}X\hspace{-0.2mm}|\hspace{-0.2mm}Z)\hspace{-0.6mm}$ is Arimoto's conditional $\alpha$-entropy defined as:\vspace{-1ex}
\begin{equation}
    \begin{split}
    H^{A}_{\alpha}(X|Z) & = \frac{\alpha}{1-\alpha} \log \sum_{z}{p_{Z}(z)}\left(\sum_{x}{p^{\alpha}_{X|Z}(x|z)}\right)^{\hspace{-1.5mm}\frac{1}{\alpha}} \\
    & = \frac{\alpha}{1-\alpha} \log \mathbb{E}_Z\left[{\lVert{p_{X|Z}}\rVert}_{\alpha}\right].
    \end{split}
\end{equation}
Consequently, $\hspace{-0.4mm}H^{A}_{\alpha}\hspace{-0.3mm}(\hspace{-0.3mm}X\hspace{-0.2mm}|\hspace{-0.2mm}Z)\hspace{-0.6mm}$ is generalized to $H^{A}_{\alpha}(X^T|Z^T)$ as:\vspace{-1ex}
\begin{equation}
    \begin{split}
    \hspace{-0.9mm}H^{A}_{\alpha}\hspace{-0.3mm}(\hspace{-0.4mm}X^T\hspace{-0.3mm}|\hspace{-0.3mm}Z^T\hspace{-0.3mm})\hspace{-0.9mm} & = \hspace{-0.9mm}\frac{\alpha}{1\hspace{-0.9mm}-\hspace{-0.6mm}\alpha}\hspace{-0.5mm} \log \hspace{-0.5mm}\sum_{\substack{z^T}}\hspace{-0.3mm}{p_{Z^T}\hspace{-0.3mm}(\hspace{-0.4mm}z^T\hspace{-0.2mm})}\hspace{-1.3mm}\left(\hspace{-0.7mm}\sum_{\substack{x^T}}{\hspace{-0.3mm}p^{\alpha}_{X^T\hspace{-0.3mm}|\hspace{-0.3mm}Z^T}\hspace{-0.3mm}(x^T\hspace{-0.3mm}|\hspace{-0.2mm}z^T\hspace{-0.2mm})}\hspace{-1.5mm}\right)^{\hspace{-1.5mm}\frac{1}{\alpha}} \\
    & =\hspace{-0.9mm} \frac{\alpha}{1-\alpha} \log \mathbb{E}_{Z^T}\hspace{-1mm}\left[{\lVert{p_{X^T|Z^T}}\rVert}_{\alpha}\right],
    \end{split}\vspace{-0.2ex}
    \label{H_A_T_eq}
\end{equation}
where ${p_{X^T|Z^T}} \hspace{-0.5mm} = \hspace{-0.6mm} \prod_{t=1}^{T}{p_{{X}_{t}|{X}^{t-1},Z^T}}$. Finally, the problem of finding the optimal releaser is formulated as follows:
\begin{equation} \label{rel_opt_problem_v0}
        \inf_{p_{{Z}^{T}|W^T}} I^{A}_{\alpha}(X^T;Z^T|S) \quad \text{subject to} \quad \mathcal{D}(Z^T,Y^T) \leq \epsilon,\vspace{-0.3ex}
\end{equation}
where $\epsilon \geq 0$ is a parameter to force the releaser to control the trade-off between privacy and utility. In addition, the SI term is considered in $I^{A}_{\alpha}(X^T\hspace{-0.3mm};\hspace{-0.3mm}Z^T\hspace{-0.3mm}|S)\hspace{-0.3mm}~\hspace{-0.3mm}=\hspace{-0.3mm}~\hspace{-0.3mm}H^{A}_{\alpha}(\hspace{-0.3mm}X^T\hspace{-0.3mm}|S)\hspace{-0.3mm}~\hspace{-0.3mm}-\hspace{-0.3mm}~\hspace{-0.3mm}H^{A}_{\alpha}(\hspace{-0.3mm}{X^T}\hspace{-0.3mm}|\hspace{-0.3mm}Z^T\hspace{-0.3mm},\hspace{-0.5mm}S)$ by substituting all $p_{.|Z^T}$ by $p_{.|Z^T\hspace{-0.5mm},S}$. Given the fact that $H^{A}_{\alpha}(X^T\hspace{-0.3mm}|S)$ cannot be changed by the releaser, i.e., it does not depend on ${p_{{Z}^{T}|W^T}}$, we re-formulate \eqref{rel_opt_problem_v0} as follows:\vspace{-1ex}
\begin{equation} \label{rel_opt_problem}
        \inf_{p_{{Z}^{T}|W^T}} -\frac{1}{T}H^{A}_{\alpha}(X^T|Z^T\hspace{-0.3mm},\hspace{-0.3mm}S), \quad \text{s.t.} \quad \mathcal{D}(Z^T,Y^T) \leq \epsilon,
    \vspace{-0.5ex}
\end{equation}
where the term $\frac{1}{T}$ is included for normalization purposes. 

Finding the solution for the optimization problem in \eqref{rel_opt_problem} is not generally tractable. In addition, tackling this problem requires the availability of $p_{{X}^T|Z^T\hspace{-0.7mm},S}$. Hence, the privacy-preserving framework approximates $p_{{X}^T|Z^T\hspace{-0.7mm},S}$ by using an estimator network, called adversary. The problem of estimating $p_{X^T|Z^T\hspace{-0.7mm},S}$ by $p_{\hat{X}^T|Z^T\hspace{-0.7mm},S}$ can be optimally tackled by minimizing the KL divergence between the distributions written as \cite{info-theory}:\vspace{-0.6ex}
\begin{equation} \label{KL_problem}
        \hspace{-1.1ex}\inf_{p_{\hat{X}^T\hspace{-0.2mm}|Z^T\hspace{-0.7mm},S}}\hspace{-2.5mm} \text{KL}\hspace{-0.8mm}\left(\hspace{-0.3mm}p_{{X}^T\hspace{-0.3mm}|Z^T\hspace{-0.7mm},S}||p_{\hat{X}^T\hspace{-0.3mm}|Z^T\hspace{-0.7mm},S}\hspace{-0.8mm}\right)\hspace{-0.5mm}=\hspace{-1mm}\inf_{p_{\hat{X}^T\hspace{-0.2mm}|Z^T\hspace{-0.7mm},S}} \hspace{-1.7mm} \mathbb{E}\hspace{-1mm}\left[\hspace{-0.5mm}\log\hspace{-0.5mm} \frac{p_{X^T|Z^T\hspace{-0.7mm},S}}{p_{\hat{X}^T|Z^T\hspace{-0.7mm},S}}\hspace{-0.9mm} \right]\hspace{-1.3mm},\vspace{-0.3ex}
\end{equation}
where the expectation is with respect to $p_{X^T\hspace{-0.7mm},Z^T\hspace{-0.7mm},S}$. Note that solving \eqref{KL_problem} is equivalent to minimizing the negative log-likelihood ${\mathbb{E}\left[- \log p_{\hat{X}^T|Z^T\hspace{-0.7mm},S}({X}^T|Z^T\hspace{-0.7mm},S) \right]}$. Furthermore, we try to simplify \eqref{KL_problem} by decomposing the probability distribution $p_{\hat{X}^T|Z^T\hspace{-0.7mm},S}$, leveraged from the natural characteristics of the defined privacy-preserving problem.
We denote the releaser and the adversary as $\mathcal{R}_\theta$ and $\mathcal{A}_\phi$, which are controlled by their parameters $\theta$ and $\phi$, respectively. For $t\in\{1, 2, \dots, T\}$, the releaser $\mathcal{R}_\theta$ takes observed variables, $W^t$, as its input and generates released variables represented as $Z_t$. Using $Z^t$, the adversary $\mathcal{A}_{\phi}$ aims to estimate sensitive information $x_t$ by approximating $p_{X_t|Z^t\hspace{-0.7mm},S}$ at each time $t$ as $p_{\hat{X}_t|Z^t\hspace{-0.7mm},S}$ and then solving $\hat{x}^{*}_t=\underset{\hat{x}_t \in \mathcal{X}}{\text{argmax}} \; p_{\hat{X}_t|Z^t\hspace{-0.7mm},S}(\hat{x}_t|z^t\hspace{-0.7mm},s)$. This means, while the goal of $\mathcal{A}_{\phi}$ is to estimate $X^T$ as precisely as possible based on $Z^T$, $\mathcal{R}_{\theta}$ aims to trade-off two different objectives. On the one hand, $\mathcal{R}_{\theta}$ intends to minimize the amount of information leaked about $X^T$ from ${Z}^T$, which will mislead the adversary. On the other hand, $\mathcal{R}_{\theta}$ tries to keep $Z^T$ as close as possible to $Y^T$ by limiting the distortion between $Z^T$ and $Y^T$ below a designated value. Based on these assumptions about releaser and adversary, we can conclude that the Markov chains $(X^t,Y^t) \hspace{-0.7mm} \mkv \hspace{-0.7mm} W^t \hspace{-1mm} \mkv \hspace{-1mm} Z^t \hspace{-1mm} \mkv \hspace{-1mm} \hat{X}^t$ and $\hat{X}^{t-1} \hspace{-1mm} \mkv \hspace{-1mm} Z^t,S \hspace{-1mm} \mkv \hspace{-1mm} \hat{X}^t$ hold for $t \hspace{-0.7mm} \in \hspace{-0.7mm} \{1, 2, \dots, T\}$. Therefore, $p_{\hat{X}^T|Z^T\hspace{-0.5mm},S}$ is re-formulated as:\vspace{-1.5ex}
\begin{equation} \label{prob_dist_x_hat}
    \begin{split}
        & p_{\hat{X}^{T}|Z^T\hspace{-0.7mm},S}(\hat{x}^{T}|z^T\hspace{-0.7mm},s) = \prod_{t=1}^{T}{p_{\hat{X}_{t}|\hat{X}^{t-1}\hspace{-0.3mm},Z^T\hspace{-0.7mm},S}(\hat{x}_{t}|\hat{x}^{t-1}\hspace{-0.3mm},z^T\hspace{-0.7mm},s)} \\
        & \hspace{-1mm} = \prod_{t=1}^{T}{p_{\hat{X}_{t}|Z^T\hspace{-0.7mm},S}(\hat{x}_{t}|z^T\hspace{-0.7mm},s)}
        \overset{\text{(i)}}{=} \prod_{t=1}^{T}{p_{\hat{X}_{t}|Z^t\hspace{-0.7mm},S}(\hat{x}_{t}|z^t\hspace{-0.7mm},s)}.
        \vspace{-1.5ex}
    \end{split}
\end{equation}
where (i) corresponds to the causality constraints that the problem may have. Hence, The adversary's objective in \eqref{KL_problem} can be achieved by addressing the optimization problem written as:\vspace{-1.6ex}
\begin{equation} \label{adv_causal_opt_problem}
        \inf_{p_{\hat{X}_{t}\hspace{-0.2mm}|Z^t\hspace{-0.7mm},S}} \hspace{1mm} \frac{1}{T}\sum^{T}_{t=1}{\mathbb{E}\left[- \log p_{\hat{X}_{t}|Z^t\hspace{-0.7mm},S}({X}_{t}|Z^t\hspace{-0.7mm},S) \right]},
        \vspace{-1.2ex}
\end{equation}
and the optimization problem of the releaser, defined in \eqref{rel_opt_problem}, is converted to a practical formulation as:\vspace{-0.5ex}
\begin{equation} \label{rel_opt_problem_X_hat}
        \inf_{p_{{Z}^{T}|W^T}} -\frac{1}{T}H^{A}_{\alpha}(\hat{X}^T|Z^T\hspace{-0.3mm},\hspace{-0.3mm}S), \quad \text{s.t.} \quad \mathcal{D}(Z^T,Y^T) \leq \epsilon,
        \vspace{-0.5ex}
\end{equation}
where the distribution on \eqref{prob_dist_x_hat} is used to compute $H^{A}_{\alpha}(\hat{X}^T\hspace{-0.3mm}|Z^T\hspace{-0.8mm},\hspace{-0.4mm}S)$. This optimization problem can~be tackled with the availability of $p_{\hat{X}^T\hspace{-0.2mm}|Z^T\hspace{-0.7mm},S}$, the adversary's~output.

\begin{figure}[!t]
	\centering
	\includegraphics[width=0.75\columnwidth]{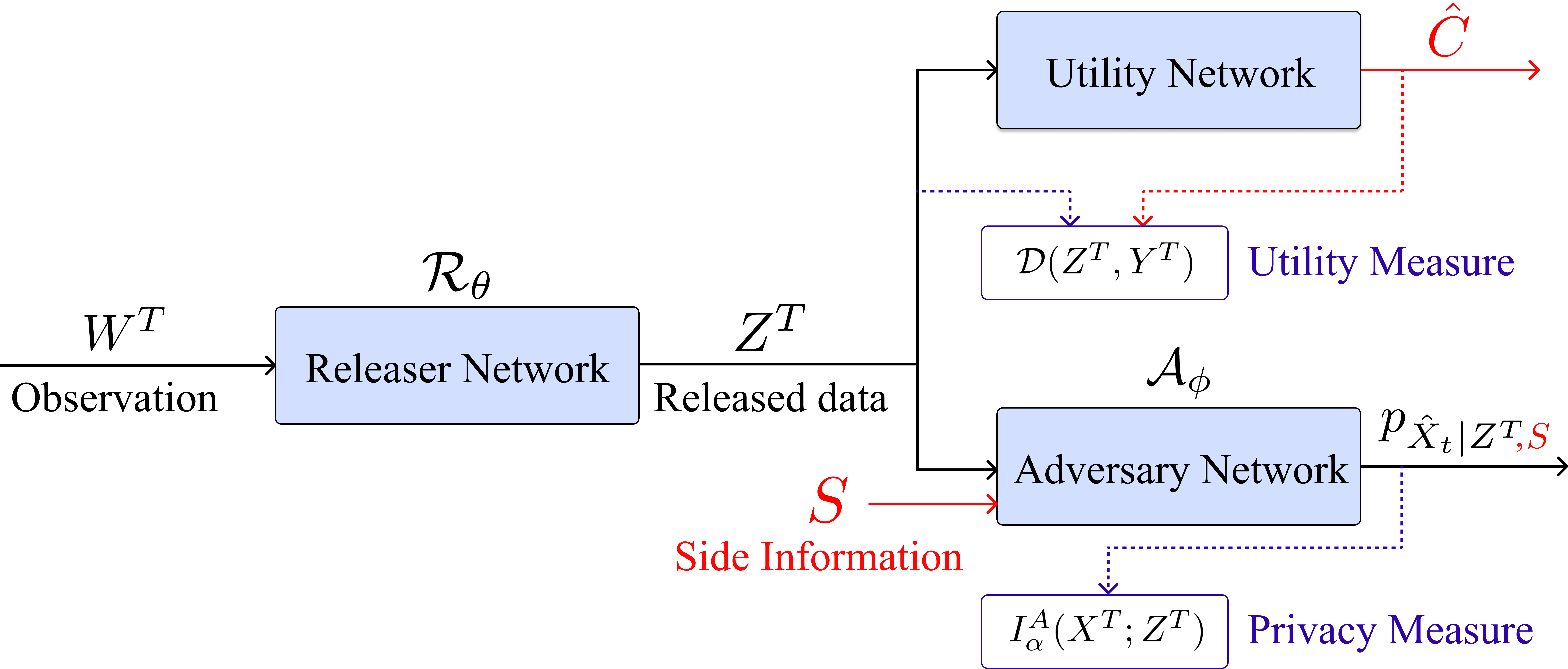}
    \vspace{-1ex}
	\caption{General privacy-preserving framework based on adversarial learning. Parts shown in red color are included as per the application and availability.}
	\vspace{-3.5ex}
	\label{fig_general_model}
\end{figure}

Based on \eqref{adv_causal_opt_problem}, $\mathcal{A}_{\phi}$ tries to maximize the quantified information between $X^T$ and $Z^T$ by minimizing KL distance between $p_{\hat{X}_{t}|Z^t}$ and $p_{{X}_{t}|Z^t}$. On the other hand, $\mathcal{R}_{\theta}$ aims to minimize $\alpha$-MI in \eqref{rel_opt_problem}. Arimoto's $\alpha$-MI is known to be a generalization for MI to measure the information shared between random variables \cite{ARIMOTO, alpha-mutual-verdu}. This suggests that the adversary's goals and the releaser's are in opposite directions. Thus, addressing \eqref{rel_opt_problem} and \eqref{adv_causal_opt_problem} can be done by a stable adversarial training procedure that uses the general modeling framework illustrated in Fig.~\ref{fig_general_model}. Two loss functions $\mathcal{L}_{\mathcal{R}}(.)$ and $\mathcal{L}_{\mathcal{A}}(.)$ are determined for $\mathcal{R}_{\theta}$ and $\mathcal{A}_{\phi}$, respectively. Using~\eqref{adv_causal_opt_problem}, $\mathcal{L}_{\mathcal{A}}({\phi})$ is written as:\vspace{-1.5ex}
\begin{equation} \label{adv_loss}
    \mathcal{L}_{\mathcal{A}}(\phi) := \frac{1}{T}\sum^{T}_{t=1}{\mathbb{E}\left[- \log p_{\hat{X}_t|Z^t\hspace{-1mm}, S}({X}_t|Z^t\hspace{-1mm},S) \right]}.\vspace{-0.6ex}
\end{equation}
As previously mentioned, \eqref{adv_loss} represents cross-entropy loss which establishes a classifier that generates $p_{\hat{X}^T|Z^T,S}$. The releaser's loss function is derived from \eqref{rel_opt_problem} as:\vspace{-1ex}
\begin{equation} \label{rel_loss}
    \mathcal{L}_{\mathcal{R}}(\theta,\phi,\omega,\alpha,\lambda) \hspace{-0.5mm} := \hspace{-0.5mm} \mathcal{D}(Z^T,Y^T) -  \frac{\lambda}{T}{H^{A}_{\alpha}(\hat{X}^T|{Z}^T\hspace{-1mm},S)}.
    \vspace{-0.8ex}
\end{equation}
The presence of $S$ in \eqref{adv_loss} and \eqref{rel_loss} depends on the availability of SI. Adjusting $\lambda\hspace{-0.5mm}\geq\hspace{-0.5mm}0$ in \eqref{rel_loss} is equivalent to changing $\epsilon$ in \eqref{rel_opt_problem}.
Considering the extreme cases, $\lambda\hspace{-0.5mm}=\hspace{-0.5mm}0$ leads the releaser to the full utility regime, meaning that $\mathcal{R}_\theta$ acts independently from $\mathcal{A}_\phi$, hence provides no privacy guarantees. For large $\lambda$ values, the term $-\frac{\lambda}{T}{H^{A}_{\alpha}(\hat{X}^T|{Z}^T\hspace{-1mm},S)}$ will be dominant in $\mathcal{L}_{\mathcal{R}}(.)$. Thus, the releaser tends to achieve full privacy, i.e., random guessing performance, by confusing the adversary totally. Moreover, $\omega$ in \eqref{rel_loss} shows the parameters that the utility network could have. Depending on the application, this network should have a complex structure or should only evaluate the specified distortion measure. Moreover, this network may generate $\hat{C}$, to which, in some applications, the distortion metric compares specific features of the useful data.

\vspace{-0.5ex}
\section{Framework and Implementation} \label{sec:model}



\subsection{Privacy-preserving framework for image data}
\label{sec:image_data}

\begin{figure}
    \centering
    \includegraphics[width=\columnwidth]{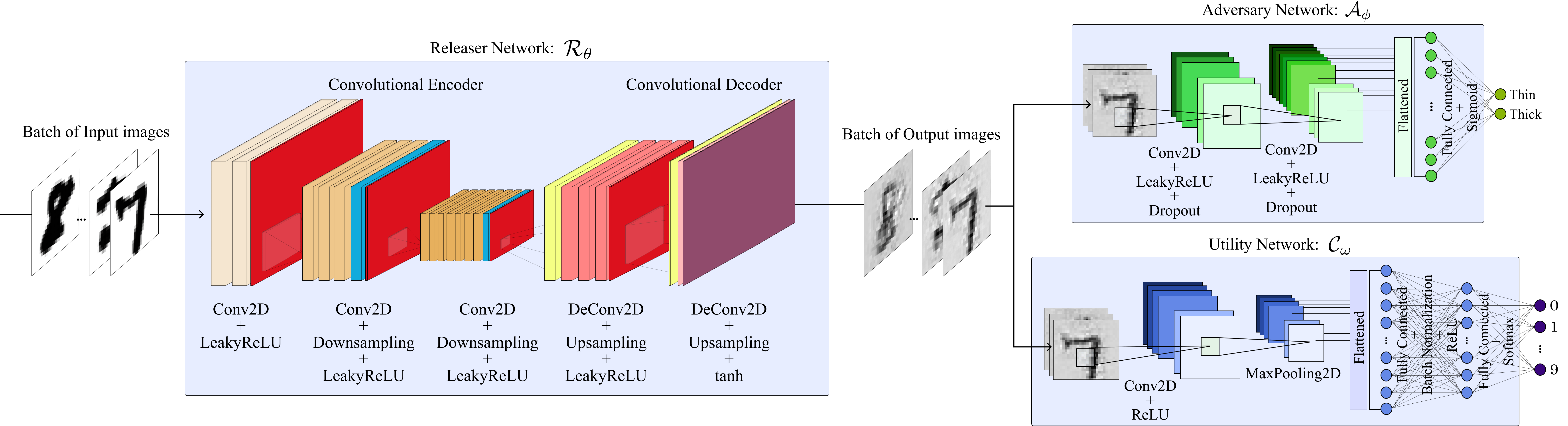}
    \vspace{-4ex}
    \caption{Privacy-preserving framework for image datasets.}
    \label{fig:amnist-model}
    \vspace{-3.5ex}
\end{figure}

Convolutional neural networks (CNNs) excel in various machine learning tasks, particularly with image datasets. Hence, in this application, we decided to build the networks shown in Fig.~\ref{fig_general_model} by using CNN modules and well-known structures related to each network's task. As illustrated in Fig.~\ref{fig:amnist-model}, $\hspace{-0.6mm}Y^T\hspace{-0.6mm}$ is considered as the releaser's input (i.e., $W^T \hspace{-1.3mm}=\hspace{-0.6mm} Y^T$). An encoder-decoder approach has been employed to design $\mathcal{R}_{\theta}$, while the adversary and utility network are image classifiers. In this work, we choose a dataset of hand-written digits where the digits' thickness is considered as private information. Thus, $\mathcal{A}_{\phi}$ tries to determine whether an image shows a thick or a thin digit. On the other hand, $\mathcal{R}_{\theta}$ aims to generate an image with the same dimensions as $\hspace{-0.6mm}Y^T\hspace{-0.6mm}$ while minimizing the distortion between the generated and original image.

The distortion measure typically quantifies the difference between the network's input and output, either on an element-wise basis or through a higher-level approach. For example, while the thickness of digits is the sensitive information that we try to hide, the ability to classify the digits is of interest. Here, an element-wise measure cannot guarantee digit classification. We consider that the distortion measure consists of two parts: (i) a $p$-norm metric that quantifies the distortion happened to the input variables, written as $d_p(Z^T,Y^T) \triangleq \frac{1}{T}{\lVert Z^T-Y^T \rVert}_{p}$ for $p \geq 1$; (ii) the loss function of the utility network, which is a categorical cross-entropy loss for an image classifier that recognizes digits. The first part of the distortion measure ensures that the released image will have element-wise similarity with the input, while the second part promotes the similarity in terms of the results of image classification.
In this application, We consider $p\hspace{-0.5mm}=\hspace{-0.5mm}1$ as the first part of the distortion measure, and the second part comes from the utility network, $\hspace{-0.3mm}\mathcal{C}_{\omega}$, written as:\vspace{-0.5ex}
\begin{equation}
    d_{\mathcal{C}}(Z^T,Y^T) \triangleq \mathcal{L}_{\mathcal{C}}(\omega)= {\mathbb{E}\left[- \log p_{\hat{C}|Z^T}({C}|Z^T,S) \right]},
    \vspace{-0.5ex}
\end{equation}
where $C$ represents particular utility features (e.g., the labels of hand-written digits images), and $\hat{C}$ is the utility network's output. Finally, the distortion measure is derived as:\vspace{-0.8ex}
\begin{equation}
    d_{\textsc{IMG}}(Z^T,Y^T) = d_{\mathcal{C}}(Z^T,Y^T) + \frac{1}{T}{\lVert Z^T-Y^T \rVert}_{1}
    \vspace{-0.5ex}
\end{equation}

For the model shown in Fig.~\ref{fig:amnist-model}, $\mathcal{A}_{\phi}$ has a cross-entropy loss, defined in \eqref{adv_loss}, and the loss function of $\mathcal{R}_{\theta}$ is formulated as:\vspace{-0.8ex}
\begin{equation} \label{amnist-rel_loss}
    \mathcal{L}_{\mathcal{R}}\hspace{-0.4mm}(\hspace{-0.4mm}\theta\hspace{-0.4mm},\hspace{-0.5mm}\phi\hspace{-0.1mm},\hspace{-0.2mm}\omega\hspace{-0.2mm},\hspace{-0.4mm}\alpha,\hspace{-0.5mm}\lambda\hspace{-0.2mm})\hspace{-0.9mm}:=\hspace{-0.5mm}\mathbb{E}\hspace{-0.7mm}\left\{\hspace{-0.6mm}d_{\textsc{IMG}}\hspace{-0.2mm}(\hspace{-0.4mm}Z^T\hspace{-0.9mm},\hspace{-0.3mm}Y^T\hspace{-0.1mm})\hspace{-0.8mm}\right\}-\frac{\lambda}{T}{H^{A}_{\alpha}\hspace{-0.2mm}(\hspace{-0.4mm}\hat{X}^T\hspace{-0.1mm}|\hspace{-0.1mm}{Z}^T\hspace{-1.2mm},S)}.
    \vspace{-0.5ex}
\end{equation}

The training process for the data releaser model of this work has multiple stages. At every training iteration, $\mathcal{A}_{\phi}$ is trained $k$ times, while $\mathcal{R}_{\theta}$ is only trained once per iteration.
The choice of $k$ is crucial as it affects the adversary's strength~\cite{gan}.
Algorithm~\ref{alg:image} provides a detailed training procedure.
After the training phase, a distinct network, called attacker, is considered for the test phase. This network is trained with the released data and will test the privacy achieved by the model. This network plays the role of a real-world attacker, which has an approximately similar structure to $\mathcal{A}_{\phi}$ and tries to infer sensitive information from released data.\vspace{-2ex}
\begin{algorithm}
 \caption{Training of privacy-preserving framework.\\
 \textbf{Hyperparameters:} {Batch size $B$, Adversary training steps $k$.}}
 \label{alg:image}
 \begin{algorithmic}[1]
\FOR {number of iterations}
  \FOR {$k$ steps}
  \STATE {Sample $\{y^{(b)T}\hspace{-1.4mm} ,\hspace{-0.3mm} x^{(b)T}\}^{B}_{b=1}$ to create $\{w^{(b)T}\}^{B}_{b=1}$.}
  \STATE {Generate $\{z^{(b)T}\}^{B}_{b=1}$ by using $\{w^{(b)T}\}^{B}_{b=1}$ and $\mathcal{R}_\theta$}.
  \STATE {Compute gradient of $\mathcal{L}_{\mathcal{A}}(\phi)$, approximated with $\{z^{(b)T}\}^{B}_{b=1}$, or $\{z^{(b)T}\hspace{-0.5mm},\hspace{-0.3mm} s^{(b)T}\}^{B}_{b=1}$ when SI is available.}
  \STATE {Update $\phi$ based on the gradient of $\mathcal{L}_{\mathcal{A}}(\phi)$.}
  \STATE {If available, compute gradient of the utility network's loss and update $\omega$ based on the gradient.}
  \ENDFOR
  \STATE {Sample $\{y^{(b)T}\hspace{-1.4mm} ,\hspace{-0.3mm} x^{(b)T}\}^{B}_{b=1}$ to create $\{w^{(b)T}\}^{B}_{b=1}$.}
  \STATE {Compute gradient of $\mathcal{L}_{\mathcal{R}}(\theta)$, approximated with $\{w^{(b)T}\}^{B}_{b=1}$, and update $\theta$ based on the gradient.}
  \ENDFOR
\end{algorithmic}
\end{algorithm}
\vspace{-3ex}
\subsection{Privacy-preserving framework for time-series data}
\label{sec:timeseries_data}
Our second example deals with time-series data. The most important feature of time series is the correlation of data points over time. In order to extract this feature, we use Long Short-Term Memory (LSTM) modules to build releaser and adversary networks of the general model shown in Fig.~\ref{fig_general_model}.
We form $W^T$ by concatenating $Y^T$ and $X^T$.
This study focuses on time-series applications where utility is defined as the similarity between released data and actual observations, such as smart grid applications~\cite{hadi-journal-alpha1-ECOwoSI}.
Hence, a $p$-norm distortion is sufficient to compare the input and output of the releaser. Therefore, we choose $d_{\textsc{TS}}(Z^T,Y^T) \hspace{-0.5mm}=\hspace{-0.5mm} \frac{1}{T}{\lVert Z^T-Y^T \rVert}_{p=2}$ as the distortion measure in this application, and there is no need to have a complex utility network. Finally, the loss function for $\mathcal{A}_{\phi}$ is the same as \eqref{adv_loss}, and, for releaser $\mathcal{R}_{\theta}$, it becomes:\vspace{-1.2ex}
\begin{equation} \label{ts-rel_loss}
    \mathcal{L}_{\mathcal{R}}\hspace{-0.3mm}(\theta,\phi,\alpha,\lambda) \hspace{-1mm} := \hspace{-0.5mm} \mathbb{E}\hspace{-0.5mm}\left\{\hspace{-0.4mm}d_{\textsc{TS}}(Z^T,Y^T)\hspace{-0.4mm}\right\} - \frac{\lambda}{T}{H^{A}_{\alpha}\hspace{-0.2mm}(\hat{X}^T|{Z}^T\hspace{-1.2mm},S)}.
    \vspace{-0.6ex}
\end{equation}

The training procedure of this framework is available by adjusting Algorithm~\ref{alg:image} based on time-series properties. Similar to section~\ref{sec:image_data}, a distinct attacker evaluates the privacy attained by the model.

\vspace{-0.5ex}
\section{Results and Discussion} \label{sec:results}
\vspace{-0.5ex}
\subsection{Datasets description} \label{sec:datasets}
\vspace{-0.2ex}
\subsubsection{Annotated MNIST (AMNIST) dataset} \label{sec:AnnMNIST}

We use the well-known MNIST dataset \cite{mnist_data_ref} and modify it by adding a label of thickness level to images using the method provided in \cite{ann_mnist_ref}.
In \cite{ann_mnist_ref}, authors have defined mathematical formulas with different parameters for each digit. Therefore, the digit thickness in a particular sample image can be classified into thick, normal, or thin. We customized the provided code in \cite{ann_mnist_ref} to label all training and testing images, and we excluded those images with a normal thickness for computational simplicity. We ended up with 28,568 training
and 4,681 testing samples.

\subsubsection{ECO dataset} \label{sec:ECO}
The Electricity Consumption and Occupancy (ECO) dataset~\cite{eco_data_ref2}
contains power consumption data of 6 households and their ground truth occupancy information. Since, in this work, the consumption data and occupancy labels are re-sampled at every hour, ECO would be considered a time-series dataset with $T=24$. Here, the power consumption represents the utility feature $Y_t$, while the household occupancy is the private information {$X_t$}. We partitioned data into 8980 training and 2245 testing time-series sequences. Moreover, week's day and month are possible SI available in ECO that can be concatenated to training and testing samples.

\vspace{-1.1ex}
\subsection{Metrics} \label{sec:metrics}

We choose Normalized Error (NE), i.e., Normalized Mean Squared Error (NMSE), to evaluate the distortion between $Y^T$ and $Z^T$. We employ balanced accuracy to compare models' performance. This metric is used instead of accuracy to mitigate the unbalanced data effects. Henceforward, we use the word {\textit{accuracy}} to refer to \textit{balanced accuracy}, for brevity.

\vspace{-1ex}
\subsection{Tunable privacy measure for AMNIST dataset} \label{sec:amnist_results}

The effects of the proposed tunable privacy measure are evaluated by performing an experiment using the modified AMNIST dataset and the proposed framework for image data. We selected $\alpha\hspace{-0.5mm}=\hspace{-0.5mm}1$ (equivalent to MI) and explored the intervals $(0,\hspace{-0.5mm}1)$ and $(1,\hspace{-0.5mm}\infty)$ to examine the model's performance by varying $\alpha$. The outcomes revealed that models with $\alpha\hspace{-0.5mm}<\hspace{-0.5mm}1$ exhibit analogous behavior, with only insignificant differences. The same phenomenon holds for $\alpha\hspace{-0.5mm}>\hspace{-0.5mm}1$.
Thus, the following values are considered for the experiments in this work: $\alpha\hspace{-0.5mm}=\hspace{-0.5mm}0.9, 1,$ $3$.

The details of the layers used in the framework are demonstrated in Fig.~\ref{fig:amnist-model}. The hyperparameters in Algorithm \ref{alg:image} are set to $B=256$ and $k=3$. Here, full privacy is achieved when the attacker cannot guess better than 50\% since we consider that the thickness has two possible values. The attacker's structure is similar to the adversary model described in Fig.~\ref{fig:amnist-model}. 

\begin{figure}[t]
    \centering
    \includegraphics[width=.78\columnwidth]{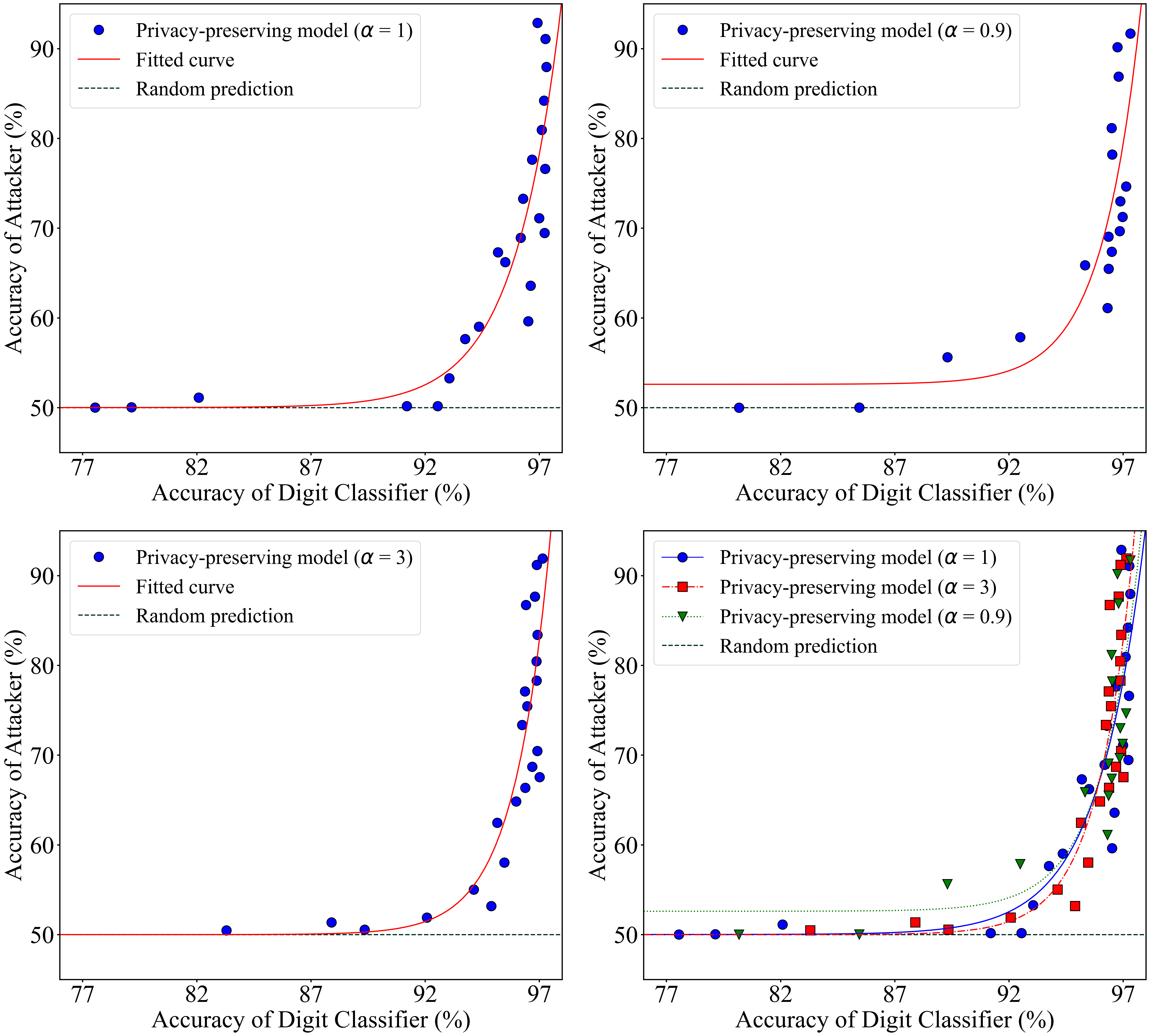}
    \vspace{-1.5ex}
    \caption{Privacy-utility trade-off for digits' thickness inference in models with different privacy measures (tuned by changing $\alpha$). The fitted curves are exponential functions and are shown only for illustration purposes.}
    \label{fig:AMNIST_privacy_utility_tradeoff_connected}
    \vspace{-4ex}
\end{figure}

In Fig.~\ref{fig:AMNIST_privacy_utility_tradeoff_connected}, the PUT for digits' thickness inference is shown.
Note that by using the original images, $Y^T$, a model can classify the digits and predict their thicknesses with 97.25\% and 91.50\% accuracy, respectively.
As illustrated in Fig.~\ref{fig:AMNIST_privacy_utility_tradeoff_connected}, for all models, the classification accuracy is almost preserved where the attacker's accuracy is around 60\%. Moreover, the classification accuracy is significantly high around the first point in the full privacy region (FPR). This result ensures achieving the essential utility goal, which is the ability to classify the released digits with high accuracy.
The behavior around edge cases is almost the same for all models, except that the model with $\alpha\hspace{-0.3mm}=\hspace{-0.3mm}0.9$ reaches the FPR with lower classification accuracy than others. The result shows the power of $\alpha\hspace{-0.3mm}=\hspace{-0.3mm}3$ while transitioning from full utility region (FTR) to the middle of the curve by reducing attacker's accuracy the most, with a slight change in digit classification. However, in the (FTR), $\alpha\hspace{-0.4mm}=\hspace{-0.4mm}1$ suggest better classification accuracy.
Notably, the model with $\alpha\hspace{-0.3mm}=\hspace{-0.3mm}0.9$ is very sensitive to small changes of $\lambda$ in \eqref{rel_loss}, which is necessary for generating points of the PUT curve. Due to this sensitivity, finding a point in the middle of the curve requires more effort than other $\alpha$ values.

\begin{figure}
    \centering
    \subfloat[$\alpha=3$\label{fig:digits_al_3.0}]
    {
    \centering
        \includegraphics[width=.9\columnwidth]{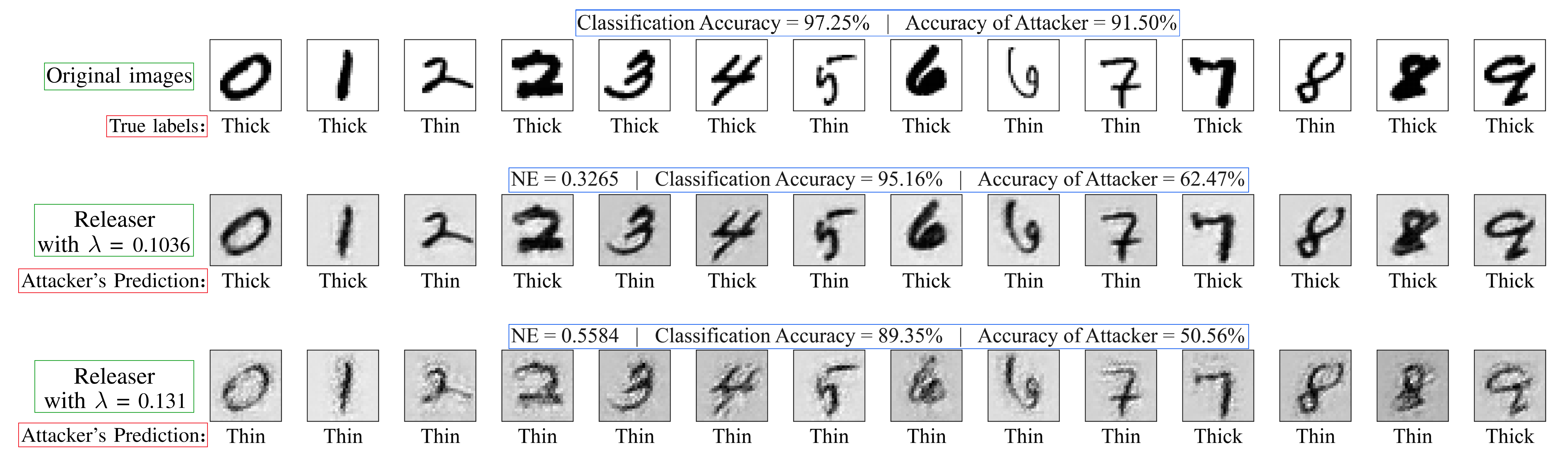}
    }
    \\
    \vspace{-1.9ex}
    \subfloat[$\alpha=1$\label{fig:digits_al_1.0}]
    {
    \centering
        \includegraphics[width=.9\columnwidth]{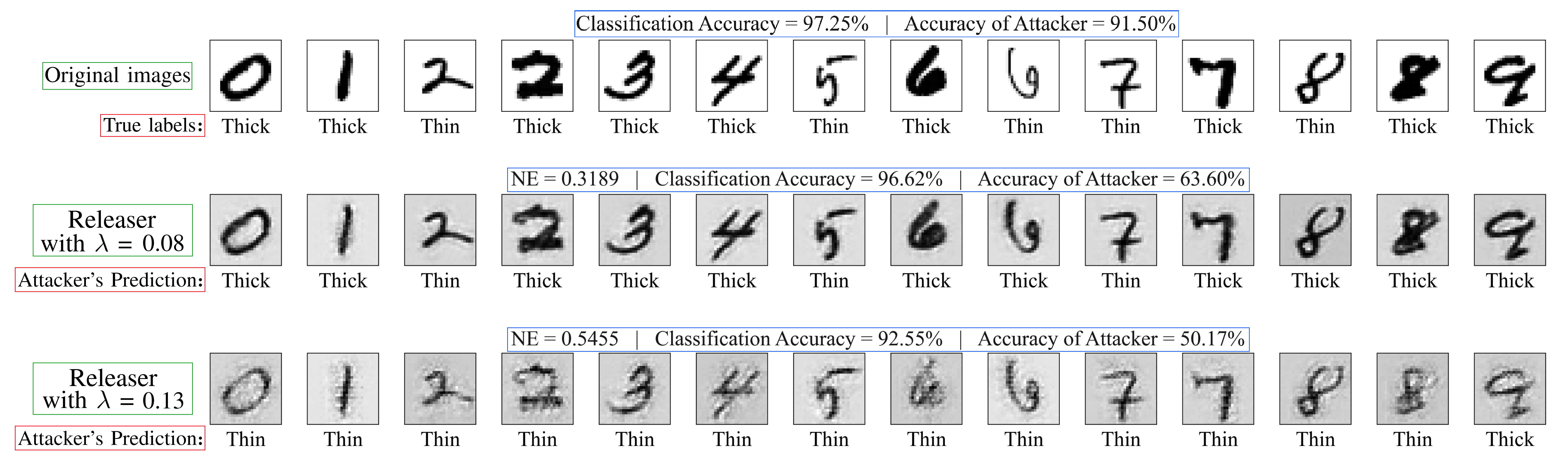}
    }
    \\
    \vspace{-1.9ex}
    \subfloat[$\alpha=0.9$\label{fig:digits_al_0.9}]
    {
    \centering
        \includegraphics[width=.9\columnwidth]{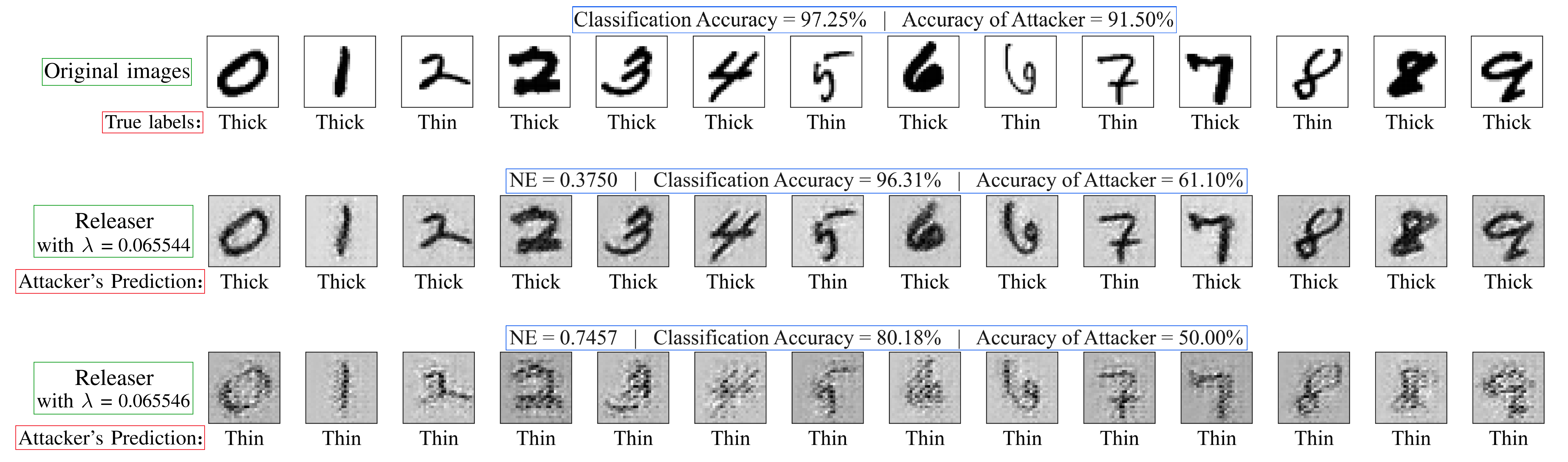}
    }
\vspace{-0.6ex}
\caption{Samples of the released images of the privacy-preserving framework for the AMNIST dataset with $\alpha=3$, $1$, and $0.9$.}
\label{fig:sample_digits}
\vspace{-2ex}
\end{figure}

Fig.~\ref{fig:sample_digits} shows examples of the released images for selected models.
For each sub-figure, we select a point in the middle of the PUT and the first point in the FPR. The results corresponding to middle of the PUT illustrate that by losing a small quantity of digit classification accuracy, the attacker's accuracy is dropped by about 30\%. Interestingly, each model's distortion has occurred differently in the full privacy examples.
These results indicate no best value of $\alpha$ for all desired operating points on the PUT. Therefore, $\alpha$ gives a degree of freedom to find a model that works best in a desired region.

We design another attacker which has gained access to the algorithm of~\cite{ann_mnist_ref}. We refer to it as the "Thickness-Computing Attacker~(TCA)." Using the algorithm, TCA can label digits as thick, normal, or thin. Since we excluded digits with normal thickness from the experiment's data, the attacker has three options for labeling digits for which the algorithm predicts normal thickness:  to assign 1) random, 2) thin, or 3) thick labels. We considered all cases for each model and reported their maximum accuracy. Some results of TCA are compared with the deep attacker~(DA) in Table~\ref{tab:amnist_attacker}. DA is stronger than TCA around the middle of the PUT; however, TCA achieves better accuracy near the FPR. Interestingly, this large gap happens for models with $\alpha\hspace{-0.3mm}=\hspace{-0.3mm}3$ and $1$ when the attackers decide to convert normal labels to thin. However, for $\alpha\hspace{-0.3mm}=\hspace{-0.3mm}0.9$, converting to thick labels is the selected approach. Since the gap is negligible in this case, we conclude that the model with $\alpha\hspace{-0.3mm}=\hspace{-0.3mm}0.9$ is more robust against different attackers than others.
\begin{table}
\scriptsize
\centering
\caption{Thickness inference results of different attackers}
\vspace{-1.5ex}
\label{tab:amnist_attacker}
\begin{tabular}{ c|c|c|c } 
 Model Parameters & DA's Accuracy & TCA's Accuracy & {Labeling }\\
 \hline
 $\alpha\hspace{-0.5mm}=\hspace{-0.5mm}3$, $\lambda\hspace{-0.5mm}=\hspace{-0.5mm}0.1036$ & $62.47$\% & $55.82$\% & Thin\\ 
 $\alpha\hspace{-0.5mm}=\hspace{-0.5mm}3$, $\lambda\hspace{-0.5mm}=\hspace{-0.5mm}0.131$ & $50.56$\% & $57.10$\% & Thin\\
 $\alpha\hspace{-0.5mm}=\hspace{-0.5mm}1$, $\lambda\hspace{-0.5mm}=\hspace{-0.5mm}0.08$ & $63.60$\% & $62.15$\% & Thin\\
 $\alpha\hspace{-0.5mm}=\hspace{-0.5mm}1$, $\lambda\hspace{-0.5mm}=\hspace{-0.5mm}0.13$ & $50.17$\% & $56.65$\% & Thin\\
 $\alpha\hspace{-0.5mm}=\hspace{-0.5mm}0.9$, $\lambda\hspace{-0.5mm}=\hspace{-0.5mm}0.065544$ & $61.10$\% & $60.28$\% & Thin\\
 $\alpha\hspace{-0.5mm}=\hspace{-0.5mm}0.9$, $\lambda\hspace{-0.5mm}=\hspace{-0.5mm}0.065546$ & $50.00$\% & $50.60$\% & Thick
\end{tabular}
\vspace{-4ex}
\end{table}

\vspace{-3ex} 
\subsection{Tunable privacy measure for ECO dataset} \label{sec:eco_results}
\vspace{-0.3ex}
Moving forward with ECO dataset, $\alpha\hspace{-0.3mm}=\hspace{-0.3mm}3,\hspace{-0.3mm}1,\hspace{-0.3mm}0.9$ are selected based on the discussed reason in section~\ref{sec:amnist_results}. The general framework illustrated in Fig.~\ref{fig_general_model} is customized based on section~\ref{sec:timeseries_data}. In addition, an independent uniformly distributed (over $[0,1]$) noise $U^T\hspace{-0.3mm}$ is integrated into $W^T\hspace{-0.3mm}$ beside $Y^T\hspace{-0.3mm}$ and $X^T$ to randomize $Z^T$. It is seen to be helpful in practical applications where an adversarial framework's input consists of noise~\cite{seed-noise}. The releaser network consists of 4 LSTM layers, each with 64 cells, and the adversary network is formed of 3 LSTM layers, each with 32 cells. The distinct attacker has the same structure as the adversary. The hyperparameters indicated in Algorithm~\ref{alg:image} are set to $B\hspace{-0.5mm}=\hspace{-0.5mm}128$, $k\hspace{-0.5mm}=\hspace{-0.5mm}4$. As discussed in section~\ref{sec:ECO}, household occupancy is private information. Thus, the corresponding attack accuracy of the FPR is 50\%. Notably, an attacker can predict household occupancy from the actual power consumption with more than 90\% accuracy.

\begin{figure}[t]
    \centering
    \includegraphics[width=.79\columnwidth]{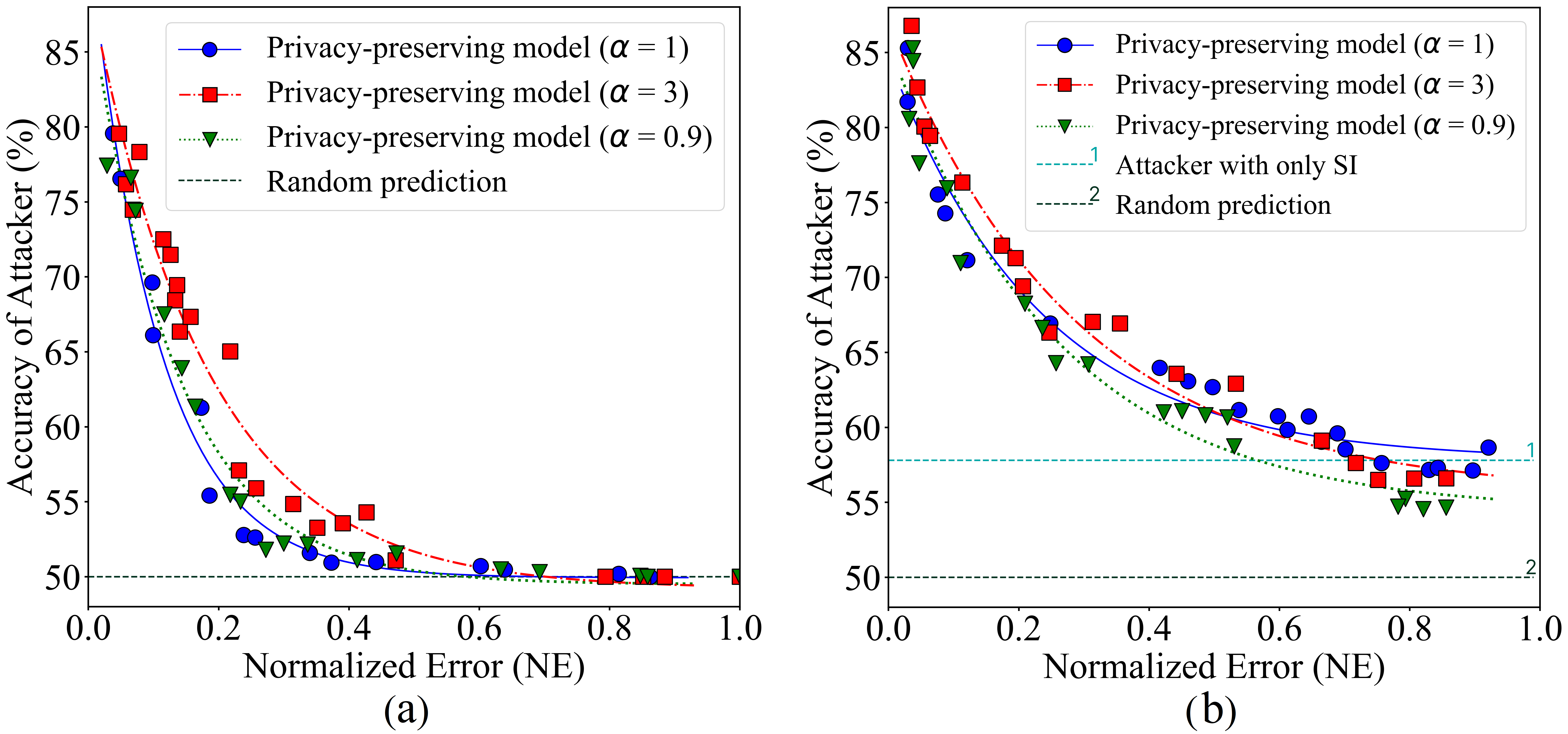}
    \vspace{-2.1ex}
    \caption{Privacy-utility trade-off for house occupancy inference in models with different privacy measures. (a) without SI, (b) SI is available to the attacker.}
    \label{fig:ECO_privacy_utility_tradeoff}
    \vspace{-2.7ex}
\end{figure}

\begin{figure}[t]
    \centering
    \includegraphics[width=.78\columnwidth]{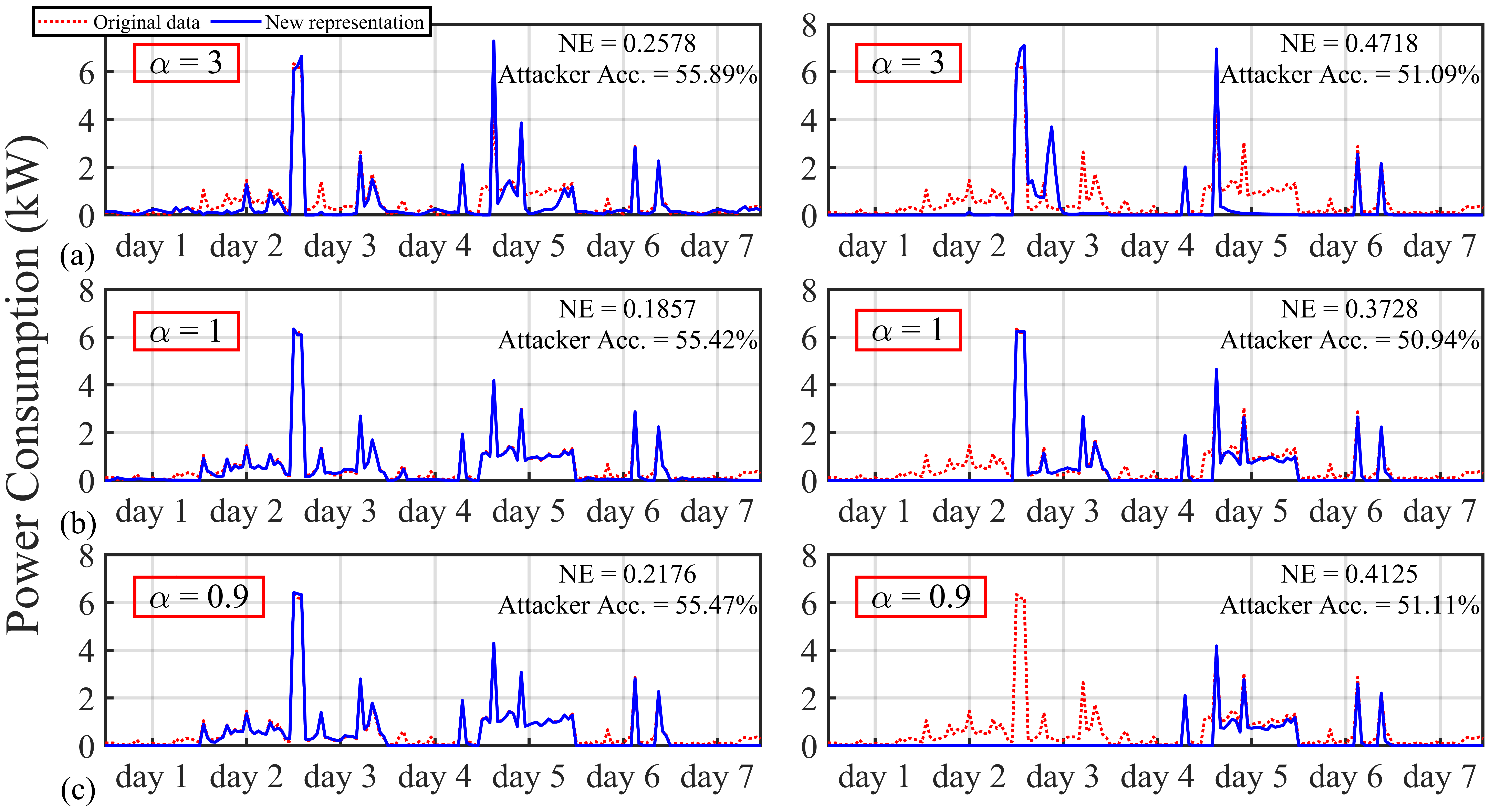}
    \vspace{-1.6ex}
    \caption{Samples of the released power consumption modified by privacy-preserving framework for time-series datasets with $\alpha=3, 1,$ and $0.9$.}
    \label{fig:ECO_data_representation}
    \vspace{-3.5ex}
\end{figure}

The PUT for house occupancy inference is available in Fig.~\ref{fig:ECO_privacy_utility_tradeoff}a. In \cite{hadi-journal-alpha1-ECOwoSI}, a similar experiment is investigated where MI is the privacy measure. Here, around FTR and FPR, all models accomplish almost the same trade-off. However, the model with $\alpha\hspace{-0.5mm}=\hspace{-0.5mm}1$ performs best in the middle of the PUT. Fig.~\ref{fig:ECO_data_representation} shows 7-day-long samples from modified power consumption signals. In this figure, two operating points are selected for each $\alpha$. The models corresponding to the left side of Fig.~\ref{fig:ECO_data_representation} preserve most of the original data (NE is less than 0.26 in the worst case), while the attacker's accuracy is dropped by more than 26\%. In addition, different distortion patterns can be realized on the right side of the figure for different $\alpha$ values.

Another experiment is designed with ECO for a situation where the SI discussed in section~\ref{sec:ECO} is available to an attacker. Fig.~\ref{fig:ECO_privacy_utility_tradeoff}b shows the PUT for selected $\alpha$ values. Similar work is conducted in \cite{side-info-hadi}, where MI is the privacy measure.
In \cite{side-info-hadi}, an attacker trained and tested with only SI achieves an accuracy of 57.8\%, concluding that the attacker is not completely confused even by signals with large distortion.
In Fig.~\ref{fig:ECO_privacy_utility_tradeoff}b, the model with $\alpha\hspace{-0.5mm}=\hspace{-0.5mm}1$ attains the attacker's accuracy of 57.8\% on large NE, while surprisingly, the model with $\alpha\hspace{-0.5mm}=\hspace{-0.5mm}0.9$ maintains the accuracy of 55.2\%. In addition, The baseline for the model with $\alpha\hspace{-0.5mm}=\hspace{-0.5mm}3$ is 56.8\%.
These results suggest better performance than \cite{side-info-hadi} in preserving sensitive information of a highly distorted signal when SI is available to the attacker.



\vspace{-0.6ex}
\section{Conclusion} \label{sec:conclusion}
\vspace{-0.6ex}
This research proposes a general privacy-preserving data-sharing model that allows for tunable privacy measures, particularly leveraging $\alpha$-Mutual Information. A key finding of the research is the influential role of the $\alpha$ parameter, which can be adjusted to balance privacy and utility in various scenarios. Experimental tests, using an image dataset of handwritten digits and a time-series sequence of power consumption measurements, revealed that tuning $\alpha$ allows for tailored data-sharing frameworks, with signals released per specific features of interest. The research also considered scenarios where attackers have access to correlated SI. The results indicated that fine-tuning of the privacy measure should consider not just the PUT, but also the model's resilience against SI. Lastly, in addition to the generic attacker of the framework, an arithmetic attacker was considered in relation to the AMNIST dataset used in this work. The results highlighted that certain models (with different privacy metrics) may be more or less successful at concealing sensitive information, depending on whether the attacker knows private information's pattern in the actual data.

\vspace{-1.2ex}

\bibliographystyle{ieeetr}
\bibliography{HREF}

\end{document}